\documentclass[10pt,twocolumn,a4paper]{article}   % <-- TWO COLUMN

\usepackage[T1]{fontenc}
\usepackage[utf8]{inputenc}
\usepackage[english]{babel}
\usepackage{lmodern}
\usepackage{microtype}

\usepackage{setspace}
\makeatletter
\if@twocolumn
  \RequirePackage[a4paper,left=0.72in,right=0.72in,top=0.95in,
                  bottom=1.05in,columnsep=0.28in]{geometry}
\else
  \RequirePackage[a4paper,left=1.15in,right=1.15in,top=1.15in,
                  bottom=1.25in]{geometry}
\fi
\makeatother

\usepackage{amsmath,amssymb}
\usepackage{booktabs}
\usepackage{multirow}
\usepackage{array}
\usepackage{graphicx}
\usepackage{caption}
\usepackage[table]{xcolor}
\usepackage{float}

\usepackage[small,bf]{titlesec}

\usepackage[hidelinks,breaklinks]{hyperref}
\usepackage{natbib}
\newcommand{\model}{\textsc{MoganColBERT-TR}}
\newcommand{\dense}{\textsc{MoganBERT-embed}}
\newcommand{\backbone}{\textsc{MoganBERT-TR}}
\newcommand{\maxsim}{\textsc{MaxSim}}
\newcommand{\TurkColBERT}{\textsc{TurkColBERT}}
\newcommand{\code}[1]{\texttt{\small #1}}
\newcommand{\best}[1]{\textbf{#1}}
\newcommand{\ours}[1]{\textbf{#1}}

\makeatletter
\if@twocolumn
  \newenvironment{restable}{\@nameuse{table*}[t]}{\@nameuse{endtable*}}
  \newcommand{\basslikboyut}{\Large}
\else
  \newenvironment{restable}{\@nameuse{table}[H]}{\@nameuse{endtable}}
  \newcommand{\basslikboyut}{\LARGE}
\fi
\makeatother

\title{\bfseries\basslikboyut \model: A Late-Interaction Multi-Vector\\[0.15em]
       Retrieval Model for Turkish}

\author{
Furkan Yılmaz \\ \texttt{furkanyl509@gmail.com}
\and
Habibe Aleyna Taşdemir \\ \texttt{aleynattasdemir@gmail.com}
\and
Muhammed Faruk Gözay \\ \texttt{gozayfaruk@gmail.com}
}
\date{}

\newcommand{\statusline}{%
  \begin{center}\small\itshape
  Preprint. Model weights will be released at
  \url{https://huggingface.co/moganai}.
  \end{center}}

\newcommand{\abstracttext}{%
We previously reported a ModernBERT encoder trained from scratch for Turkish
(\backbone) and a single-vector embedding model built on top of it (\dense).
This work introduces the third model in that lineage: \model, a multi-vector
retrieval model that, instead of compressing a query or a document into a single
vector, represents it at the token level through a $768\!\rightarrow\!128$
projection and scores it with \maxsim{} late interaction. The model is not
trained from scratch: the embedding model's encoder is taken as the starting
point and adapted to the ColBERT objective with a single-epoch distillation
phase. Training data is produced from two sources title$\rightarrow$passage
pairs carved out of our own pretraining corpus in the character domain and at
sentence boundaries, and two Turkish question-based retrieval sets and is
distilled from the soft scores of a cross-encoder teacher
(\code{bge-reranker-v2-m3}) over one positive and seven mined negatives. We show
that in hard negative mining, rank-based skipping alone is insufficient and must
be combined with a group mask and a cosine ceiling. Evaluation is carried out
with the official pipeline of \TurkColBERT{}, a benchmark built for Turkish
late-interaction retrieval (PLAID index, exact \maxsim), on five Turkish BEIR
datasets; none of them appears in our training pool, so all five results are
clean zero-shot. With 148.9M parameters, \model{} reaches an
\textbf{overall score of 37.36} (35.53 nDCG@100, 31.81 nDCG@10) averaged over the
five datasets and finishes second among the five models compared: it outperforms
the twice-as-large \mbox{ColmmBERT-base-TR} on four of five datasets and by
$+3.05$ overall, and the benchmark's largest model by $+12.30$. The gap to the
leading model (mLateOn) is concentrated on ArguAna-TR, the dataset with by far
the longest queries.}

\begin{document}

\makeatletter
\if@twocolumn
  % two column: title and abstract span the full page width
  \twocolumn[\begin{@twocolumnfalse}
    \maketitle
    \begin{quote}\begin{abstract}\abstracttext\end{abstract}
    \vspace{0.2em}\statusline\end{quote}
    \vspace{1.4em}
  \end{@twocolumnfalse}]
\else
  \maketitle
  \begin{abstract}\abstracttext\end{abstract}
  \vspace{0.3em}\statusline
\fi
\makeatother

% =====================================================================
\section{Introduction}

The dominant design in retrieval encodes the query and the document
independently into a single dense vector each
\citep{karpukhin2020dpr,reimers2019sbert}. The cost of this design is low
document vectors are precomputed and search is a single inner product but
the information bottleneck is severe: the entire meaning of a variable-length
document is compressed into one fixed-size point. Cross-encoders remove that
bottleneck \citep{nogueira2019passage} but require a full forward pass per
query--document pair, which makes them unusable at corpus scale
\citep{lin2021pretrained}.

ColBERT \citep{khattab2020colbert} proposes a middle ground between the two
extremes: the query and the document are encoded separately and in advance (as
in a bi-encoder), but the representation is kept at the token level and the
score is computed by matching each query token with the document token most
similar to it (\maxsim). Because the interaction is deferred to a \emph{late}
stage, after encoding, the document side remains precomputable. Later work
matured this design on both the quality and the serving-cost side
\citep{santhanam2022colbertv2,santhanam2022plaid,lee2023xtr,dhulipala2024muvera}.

There is a specific reason why this design is of particular interest for
Turkish. Turkish is agglutinative; a single word carries all of the inflectional
and derivational information that English distributes across separate words
\citep{oflazer2018turkish}.
Our own tokenizer statistics show this directly: 95\% of word \emph{types} are
split into multiple pieces, yet only 41\% of the actual token stream comes from
multi-piece words. In other words, discriminative content words are
systematically split into several tokens, and the signal those pieces carry gets
blended into a single vector during mean pooling. Late interaction keeps the
pieces apart until scoring time. That a monolingual encoder can outperform its
multilingual counterparts has already been demonstrated for Japanese by the
JaColBERT line \citep{clavie2023jacolbert,clavie2024jacolbertv25}; this work
asks the same question for Turkish.

This paper documents the third model built on top of the two we reported
earlier:

\begin{enumerate}\setlength{\itemsep}{0.15em}
\item \backbone{} a 149.4M-parameter ModernBERT architecture trained from
      scratch for Turkish, pretrained on 237.3B tokens with a CLM$\rightarrow$MLM
      curriculum \citep{warner2024modernbert,moganai2026}, and the strongest of
      the Turkish ModernBERTs compared on TrGLUE (\S\ref{sec:background}).
\item \dense{} a single-vector embedding model built on that encoder with
      distillation + NCE + GOR phases; 68.30 overall on MTEB(Turkish).
\item \model{} (this work) the multi-vector retrieval model of the same
      lineage.
\end{enumerate}

Our contributions are: (i) a ColBERT model for Turkish derived from our own
pretraining corpus rather than fine-tuned from a foreign base second place on
\TurkColBERT{} with 148.9M parameters, ahead of models twice its size; (ii) a
three-way filter for hard negative mining that shows rank-based skipping to be
insufficient under Turkish corpus conditions and corrects it with a group mask
and a cosine ceiling; (iii) a reproducible data pipeline that generates ColBERT
training data from a pretraining corpus without decoding and by splitting at
sentence boundaries.

% =====================================================================
\section{Background: MoganBERT-TR and the Single-Vector Embedding Model}
\label{sec:background}

This section summarizes the two preceding models only to the level required to
read the results of this work; the detailed rationale for each process is given
in the earlier report \citep{moganai2026}.

\paragraph{Data and tokenizer.}
The pretraining corpus was produced directly from the raw monthly Common Crawl
snapshots and from FineWeb's cleaned archive \citep{penedo2024fineweb},
rather than as a subset of third-party multilingual collections; the
heuristic filter chain follows the pattern established by CCNet
\citep{wenzek2020ccnet} and FineWeb. The semantic quality filter evolved from an
LLM labelling step to a Turkish BERT \citep{devlin2019bert} classifier, and
from there to a fastText
model into which that classifier's decisions over 480,000 examples were
distilled; the final filter agrees with BERT 94.4\% of the time and is
$\sim$90$\times$ faster. The tokenizer is a 50,048-piece SentencePiece
\citep{kudo2018sentencepiece} model trained with separate normalization
pipelines for Turkish and for code.

\paragraph{Pretraining.}
The architecture is ModernBERT \citep{warner2024modernbert}, a modern
derivative of the Transformer \citep{vaswani2017attention} encoder stack
(22 layers, hidden
size 768, \code{local\_attention}=128, 149.4M parameters). Training used a two-stage CLM$\rightarrow$MLM curriculum instead of pure
MLM \citet{gisserotboukhlef2025mlm} report, in a controlled ablation
over 38 models, that this biphasic schedule outperforms pure MLM under a
fixed compute budget; in a 10,000-step
ablation this choice was practically a tie on linear probes (average $+0.64$
points) while producing a 2.7--3.7$\times$ advantage on Turkish MS~MARCO
retrieval. The reason for the divergence is embedding geometry: in the pure-MLM
run a single component absorbs 28.1\% of the variance, against 11.9\% in the
curriculum run. Full pretraining was carried out on 4$\times$H100 with 237.3B
tokens per branch; during annealing the context was extended
$1024\!\rightarrow\!8192$ and the global RoPE \citep{su2024roformer}
$\theta$ scaled
$10^4\!\rightarrow\!1.6\times10^5$. The final model scores $78.41 \pm 0.32$ on
the 8-task TrGLUE average, the best among the Turkish ModernBERTs compared
(TabiBERT 77.83, ModernBERT-TR 77.64).

\paragraph{Anisotropy: the starting point of the embedding model.}
The raw pretrained model cannot be used for retrieval directly. This is the well-documented anisotropy / representation-degeneration
problem of representations trained with a language modelling objective
\citep{gao2019degeneration,ethayarajh2019contextual}. Measured
anisotropy is $\cos_{\text{raw}} = 0.9841$: the representations of two random
sentences point in almost the same direction, and cosine similarity buries the
discriminative signal in noise. In this regime zero-shot IR stalls at 0.2361.

\paragraph{The single-vector embedding model (\dense).}
It has two phases. \emph{Phase 1} is a \emph{feature} distillation with
Qwen3-8B-Embed \citep{zhang2025qwen3embedding} as the teacher; the student is projected up from 768 into the
teacher's 3072-dimensional space and a GOR (anisotropy regularization;
\citealp{zhang2017spreadout}) term is added to the loss. This single phase changes the geometry fundamentally:
$\cos_{\text{raw}}\ 0.9841 \rightarrow 0.0851$, effective rank
$117.9 \rightarrow 157.2$, zero-shot IR $0.2361 \rightarrow 0.5927$.
\emph{Phase 2} is contrastive fine-tuning with hard-negative InfoNCE
\citep{oord2018cpc}; NLI,
graded STS (with a CoSENT loss), classification labels, QA and parallel-text
signals are added to the retrieval pairs, supported by 40,737
sentiment-contrast triplets generated from scratch. The final model was obtained
by weight-averaging the two phases (a model soup;
\citealp{wortsman2022modelsoups}) and finished ahead of the Turkish models
compared on the 26-task MTEB(Turkish) \citep{muennighoff2023mteb} overall
average with 68.30
(Retrieval subset 59.58).

\paragraph{The starting point of this work.}
The gap left by the single-vector track is in the design itself: once Phase 1
has fixed the geometry, the remaining loss comes not from anisotropy but from
\emph{compression} a variable-length document is still reduced to a single
point. \model{} aims to move past that limit by keeping the representation at
the token level. One important design decision is that the encoder is not
trained from scratch but initialized from \dense{}: the isotropic,
retrieval-aligned geometry produced by the two phases is a far better starting
point for a ColBERT projection than a raw ModernBERT encoder. Since training
also updates the encoder, the two models are independent afterwards; no
"shared encoder" claim is made.

% =====================================================================
\section{Related Work}
\label{sec:related}

\subsection{From single-vector retrieval to late interaction}

The first wave of neural retrieval is the bi-encoder design, which encodes the
query and the document independently into one dense vector each: DPR
\citep{karpukhin2020dpr} showed that this setup can outperform BM25
\citep{robertson2009bm25} in open-domain question answering, and Sentence-BERT
\citep{reimers2019sbert} carried the same architecture over to sentence
embeddings. Subsequent work improved this line along two axes: large-scale
weakly-supervised pre-training (E5, \citealp{wang2022e5}; BGE-M3,
\citealp{chen2024m3}) and better-designed negative sampling (ANCE,
\citealp{xiong2021ance}; TAS-B, \citealp{hofstatter2021tasb}). At the opposite
extreme, cross-encoders that jointly encode the query and the document
\citep{nogueira2019passage} define the quality ceiling but cannot be run at
corpus scale; a comprehensive treatment of this trade-off is given by
\citet{lin2021pretrained}. On the sparse side, SPLADE
\citep{formal2021splade} instead preserves inverted-index infrastructure through
learned term expansion.

ColBERT \citep{khattab2020colbert} places \emph{late interaction} between these
two extremes: the representation is kept at the token level and the interaction
is deferred to a \maxsim{} step after encoding. The document side therefore
remains precomputable while scoring gains a level of detail approaching that of
a cross-encoder.

\subsection{The evolution of late interaction}

ColBERTv2 \citep{santhanam2022colbertv2} moved this line into production with
two changes: distillation from a cross-encoder teacher and centroid-based
residual compression. The training design of this work follows that scheme
directly. Later work focused largely on \emph{serving} cost PLAID
\citep{santhanam2022plaid} reduced latency through centroid pruning, XTR
\citep{lee2023xtr} simplified the gathering step by rethinking the role of token
retrieval, and MUVERA \citep{dhulipala2024muvera} reduced multi-vector search to
a fixed-dimensional single-vector search compatible with existing ANN
infrastructure. PyLate \citep{chaffin2025pylate} is an open training and
inference library built on sentence-transformers \citep{reimers2019sbert}; the
training pipeline of this work runs on PyLate.

\subsection{Multilingual and monolingual late interaction}

Two opposing strategies have been followed in carrying late interaction beyond
English. The first is a \emph{multilingual base}: Jina-ColBERT-v2
\citep{jha2024jinacolbertv2} trains an encoder of the XLM-R
\citep{conneau2020xlmr} lineage on weakly-supervised multilingual pairs with
cross-encoder distillation; ColBERT-XM \citep{louis2024colbertxm} targets
zero-shot transfer from a single high-resource language to unseen languages
through modular adapters; and mLateOn \citep{sourty2026lateon} continues the
same approach on an mmBERT \citep{marone2025mmbert} base. The second is a
\emph{monolingual encoder}: JaColBERT \citep{clavie2023jacolbert} and its
successor JaColBERTv2.5 \citep{clavie2024jacolbertv25} showed that a monolingual
Japanese encoder can outperform its multilingual counterparts even under a
constrained compute budget. \model{} belongs to this second lineage and is, to
our knowledge, its first instance for Turkish: the encoder is not a fine-tune of
a foreign model but a ModernBERT \citep{warner2024modernbert} trained from
scratch on an entirely Turkish corpus \citep{moganai2026}.

\subsection{Turkish retrieval and the evaluation ground}

The common encoder bases for Turkish are BERTurk \citep{schweter2020berturk}
and, more recently, models that moved to the ModernBERT architecture such as
TabiBERT and ModernBERT-TR. The agglutinative structure of Turkish a single
word carrying the inflectional and derivational information that English
distributes across separate words \citep{oflazer2018turkish} makes subword
fragmentation a first-order design variable for retrieval.

On the evaluation side, BEIR \citep{thakur2021beir} standardized the zero-shot
protocol and MTEB \citep{muennighoff2023mteb} multi-task embedding evaluation;
multilingual extensions arrived with MIRACL \citep{zhang2023miracl} and mMARCO
\citep{bonifacio2022mmarco}. Turkish retrieval is largely measured through the
Turkish subset of MTEB; however, MTEB's retrieval runner assumes one vector per
document and cannot score a late-interaction model. \TurkColBERT{}
\citep{ezerceli2025turkcolbert} was built to fill exactly that gap: on the
Turkish versions of five BEIR datasets, with PLAID and MUVERA indexing paths, it
compares both dense and late-interaction models across the 0.2M--600M parameter
range, and it is the measurement ground of this work.

\subsection{Hard negative mining and distillation}

In dense retrieval, negative selection is one of the single most decisive data
choices for quality; ANCE \citep{xiong2021ance} refreshes negatives from the
index throughout training, while TAS-B \citep{hofstatter2021tasb} composes
batches according to topic balance. The \emph{false negative} problem a
mined candidate that is in fact relevant was addressed explicitly by
RocketQA \citep{qu2021rocketqa} and filtered with a cross-encoder. On the
distillation side, the general framework of \citet{hinton2015distilling} was
carried to retrieval models by \citet{hofstatter2020distill}, and ColBERTv2
\citep{santhanam2022colbertv2} and Jina-ColBERT-v2 \citep{jha2024jinacolbertv2}
made it standard in late interaction; multilingual rerankers (e.g.\
\citealp{chen2024m3}) are widely used as teachers.

Our setup has an additional source of false negatives that is not discussed in
the literature: because training pairs are produced by \emph{splitting} corpus
documents into passages, a second passage carved from the same document
naturally ranks near the top for the same title and is not removed by a naive
"skip the first $k$" rule (\S\ref{sec:negatives}).

% =====================================================================
\section{Model}
\label{sec:model}

\subsection{Architecture}

The mean pooling layer of the \dense{} encoder is removed and replaced by a
bias-free linear $768 \rightarrow 128$ projection applied to every token. The
dimensionality is the established setting of ColBERTv2
\citep{santhanam2022colbertv2} and governs the trade-off between index size and
representational capacity. For a
document $d$, the representation is a matrix of $L_d$ tokens:
\[
E_d = \big[\, \hat{e}_{d,1},\ \dots,\ \hat{e}_{d,L_d} \,\big]
\in \mathbb{R}^{L_d \times 128},
\qquad
\hat{e} = \frac{W e}{\lVert W e \rVert_2}.
\]
The same encoder and the same projection are used for a query $q$. The score is
the standard \maxsim{} sum of \citet{khattab2020colbert}:
\[
S(q,d) \;=\; \sum_{i=1}^{L_q} \max_{1 \le j \le L_d}\
\hat{e}_{q,i}^{\top}\, \hat{e}_{d,j}.
\]
Placing the sum on the query side and the maximum on the document side is
asymmetric, and deliberately so: every query token looks for its counterpart in
the document, while document tokens with no counterpart in the query do not
penalize the score.

\begin{restable}
\centering
\caption{\model{} configuration.}
\label{tab:config}
\small
\begin{tabular}{ll}
\toprule
\textbf{Item} & \textbf{Value}\\
\midrule
Encoder (initialization) & \dense{} (ModernBERT, 22 layers / hidden 768)\\
Parameters               & 148.9M\\
Token embedding size     & 128 (bias-free linear projection, $\ell_2$ norm.)\\
Query length             & 32 (\code{[MASK]} expansion fills this buffer)\\
Document length          & 512 (training)\\
Query prefix             & \code{[unused0]}\\
Document prefix          & \code{[unused1]}\\
Score                    & \maxsim{}\\
Library                  & PyLate / sentence-transformers\\
\bottomrule
\end{tabular}
\end{restable}

\subsection{Query expansion and length choices}

Queries are padded to a fixed 32 tokens with \code{[MASK]} (query augmentation).
This is the mechanism from the original ColBERT design of
\citet{khattab2020colbert}, retained in later multilingual adaptations as well
\citep{jha2024jinacolbertv2,clavie2024jacolbertv25}. The masked positions
open additional matching slots for terms that the query implies but does not
state, generated by the encoder from context. The value was chosen from the
length distribution of the title and question strings in the training pool; as
we show in \S\ref{sec:discussion}, this ceiling is the model's most visible
constraint on tasks whose queries are a paragraph long.

Setting the document length to 512 is forced by the length-band choice in data
generation. Training passages are drawn from the long bands of the pretraining
corpus (250--1000 and 1000--2500 tokens); truncating a 1500-token span to 300
tokens would defeat the purpose of selecting that band. 512 also covers the
document lengths of the benchmark datasets (NFCorpus $\sim$350, SciFact
$\sim$380, SciDocs $\sim$250, ArguAna $\sim$220, FiQA $\sim$180 tokens).

% =====================================================================
\section{Training Data}
\label{sec:data}

The training pool is produced from two sources: title$\rightarrow$passage pairs
extracted from our own pretraining corpus, and two Turkish question-based
retrieval sets.

\subsection{Title $\rightarrow$ passage pairs}

The 17 shards of the pretraining corpus ($\sim$1.7~GB compressed) are scanned.
For each record the title is taken as the query and the body as the positive
document. Three filters are applied:

\begin{itemize}\setlength{\itemsep}{0.15em}
\item \textbf{Domain filter.} The \code{code} and \code{math} domains are
      dropped as measured, the body text of these domains is predominantly
      English.
\item \textbf{Band filter.} Only the 250--1000 and 1000--2500 token bands are
      kept; the short band does not carry enough content for a title--body
      relation.
\item \textbf{Title quality.} Word-count bounds, underscore/code-pattern checks
      and an all-caps filter. The minimum word count is lowered to 1 for
      Wikipedia two-word titles are legitimate in that source.
\end{itemize}

\paragraph{Passage splitting without decoding.}
Bodies exceeding 512 tokens are split. The critical implementation detail is
\emph{not} to cut the token list and \code{decode} it: decoding corrupts a
character when the cut falls in the middle of a multi-byte one. Instead we move
to the character domain via the tokenizer's offset mapping and pull the cut point
\emph{back} to the nearest sentence boundary. Because the boundary can only pull
back, we get two guarantees: no passage exceeds the token budget, and every
passage is a verbatim substring of the original text. The second is also a
testable property and was used as a verification step.

\paragraph{Passage cap per document.}
At most 2 passages are taken per document. A third passage is usually a subtopic
the title does not cover; a noisy positive is more harmful than a false negative
because it shows the model an outright wrong target. Moreover, a substantial part
of the long band is academic text, and uncapped splitting would skew the source
distribution of the pool in that direction.

\subsection{Hard negative mining}
\label{sec:negatives}

Seven negatives per query are mined in the embedding space computed with the
\dense{} encoder the established pattern of ANCE \citep{xiong2021ance} and
of ColBERTv2's self-mining \citep{santhanam2022colbertv2}. The unique documents in the pool are encoded with mean pooling
($\sim$550k $\times$ 768, fp16 $\approx$ 845~MB since this fits on a single
GPU, plain \code{torch} matrix multiplication is used instead of faiss), the top
100 candidates are retrieved per query, and three filters are applied
\emph{together}:

\begin{enumerate}\setlength{\itemsep}{0.15em}
\item \textbf{Rank-based skipping.} The first 10 candidates are skipped
      these are typically either very easy or genuinely relevant documents.
\item \textbf{Group mask.} A candidate is dropped if its group identifier equals
      the query's. This is mandatory because of passage splitting: the second
      passage carved from the same document naturally scores high for the same
      title and is not a negative.
\item \textbf{Cosine ceiling.} Candidates with cosine similarity above $0.95$
      are dropped.
\end{enumerate}

The third filter closes the case the first two miss: for some queries the number
of genuinely relevant documents exceeds 10, and those documents may carry a
different group identifier (for example two texts on the same topic from two
different sources). A fixed rank threshold cannot catch this; a similarity
ceiling can. If 7 negatives are not filled after filtering, the remainder is
completed with random documents that satisfy the group mask.

\subsection{Teacher scores}

Training uses distillation rather than a contrastive objective. Every
$(\text{query},\ \text{document})$ pair 1 positive + 7 negatives is fed
to the \code{BAAI/bge-reranker-v2-m3} \citep{chen2024m3} cross-encoder a
common teacher choice for late-interaction models
\citep{santhanam2022colbertv2,jha2024jinacolbertv2} and the resulting
score
vector becomes the soft target for the student. Scores are written to disk in
shards of 20,000; if the session drops, only the remaining shards are recomputed
rather than the mining. The rate at which the teacher places the positive above
all seven negatives is also tracked as a data-health measure; a drop in that rate
is the earliest sign that false negatives are leaking through mining.

% =====================================================================
\section{Training}
\label{sec:training}

The loss is the \code{Distillation} objective of PyLate
\citep{chaffin2025pylate}: the student's \maxsim{} score distribution over the
8 documents is aligned to the teacher's score distribution with KL divergence
the framework of \citet{hinton2015distilling} adapted to retrieval
\citep{hofstatter2020distill,santhanam2022colbertv2}. The contrastive alternative was not measured; distillation
was preferred, accepting the cost of the teacher phase.

\paragraph{Two learning rates.}
The optimizer is AdamW \citep{loshchilov2019adamw}, set up with two parameter
groups: encoder $1\times10^{-5}$, projection $1\times10^{-4}$. The rationale is direct  the projection starts
random, while the encoder has been through 237.3B tokens of pretraining and two
phases of embedding training. With a single learning rate, the noisy gradient of
the random projection in the first hundred steps propagates back into the
encoder and damages that accumulation.

\begin{restable}
\centering
\caption{Training configuration.}
\label{tab:training}
\small
\begin{tabular}{ll}
\toprule
\textbf{Item} & \textbf{Value}\\
\midrule
Loss                 & Distillation (KL, teacher: \code{bge-reranker-v2-m3})\\
Documents per query  & 1 positive + 7 hard negatives\\
Batch size           & 16\\
Epochs               & 1\\
Learning rate        & encoder $1\!\times\!10^{-5}$ / proj. $1\!\times\!10^{-4}$\\
Warmup               & 5\%\\
Weight decay         & 0.01\\
Precision            & bf16\\
Attention            & FlashAttention-2 \citep{dao2023flashattention2}\\
Hardware             & 1$\times$H100\\
Wall-clock           & $\sim$6 hours\\
\bottomrule
\end{tabular}
\end{restable}

\paragraph{Smoke test.}
Before the long run, a two-step smoke test is executed with the real Trainer, the
real collator and the real loss. This validates the entire data pipeline, the
\code{[unused0]}/\code{[unused1]}/\code{[MASK]} identifiers in the tokenizer, and
the projection parameter groups without paying the cost of the long run.

% =====================================================================
\section{Evaluation Protocol}
\label{sec:protocol}

\paragraph{Why we do not build our own protocol.}
The TabiBench protocol of the earlier work \citep{moganai2026} relies on a
fine-tuned, dense-path runner and cannot score a multi-vector model. The same constraint applies to
MTEB's standard retrieval runner it assumes one vector per document. We
therefore did not use MTEB in this work, and chose \TurkColBERT{}
\citep{ezerceli2025turkcolbert}, a benchmark built \emph{specifically for
late-interaction models}, as the measurement ground. All numbers are produced
with the benchmark's \emph{official} pipeline; no metric was reimplemented and no
setting was changed in our model's favor.

\paragraph{Tasks.}
\TurkColBERT{} consists of the Turkish versions of five BEIR datasets
\citep{thakur2021beir} (Table~\ref{tab:datasets}); the approach of producing a
multilingual retrieval collection by translation is the same as that of mMARCO
\citep{bonifacio2022mmarco}; the same pattern appears in the training mixtures
of multilingual embedding models \citep{wang2024multilinguale5}. The datasets
cover five distinct retrieval types: scientific
claim verification, argument retrieval, financial answer retrieval, citation
prediction, and domain-intensive document retrieval.

\begin{restable}
\centering
\caption{\TurkColBERT{} evaluation datasets.}
\label{tab:datasets}
\small
\begin{tabular}{llrrl}
\toprule
\textbf{Dataset} & \textbf{Domain} & \textbf{Queries} & \textbf{Corpus} & \textbf{Task type}\\
\midrule
SciFact-TR \citep{wadden2020scifact}    & Scientific claims & 1,110 &  5,180 & Fact checking\\
ArguAna-TR \citep{wachsmuth2018arguana} & Argument          &   500 & 10,000 & Argument retrieval\\
FiQA-TR \citep{maia2018fiqa}            & Finance           &   600 & 50,000 & Answer retrieval\\
SciDocs-TR \citep{cohan2020specter}     & Scientific        & 1,000 & 25,000 & Citation prediction\\
NFCorpus-TR \citep{boteva2016nfcorpus}  & Nutrition         & 3,240 &  3,630 & Document retrieval\\
\bottomrule
\end{tabular}
\end{restable}

\paragraph{Zero-shot cleanliness.}
None of the five datasets appears in the training pool. There is no intersection
between the two question-based sets used in training (\code{ret\_msmarco},
\code{ret\_tquad2}) and the benchmark datasets; the title$\rightarrow$passage
side comes entirely from our own pretraining corpus. All five reported results
are therefore clean zero-shot, and the in-domain caveat we had to accept on the
MS~MARCO side of the earlier work does not apply here.

\paragraph{Indexing: PLAID, exact \maxsim.}
\TurkColBERT{} defines two indexing paths: PLAID \citep{santhanam2022plaid}
a high-fidelity baseline with exact \maxsim{} scoring and MUVERA
\citep{dhulipala2024muvera}, a fast path performing approximate search with
fixed-dimensional encodings. This work uses \emph{PLAID only}. The rationale is
methodological: our aim is to measure the representational quality of the model,
not the speed/quality trade-off of a serving stack; introducing approximate
search would obscure how much of the reported difference comes from the model and
how much from index approximation.

\paragraph{Inference settings.}
The defaults of the official pipeline are preserved: \code{document\_length}=300,
$k$=100. This value is below the 512 we used in training; that is, the model is
measured with a document window \emph{shortened} relative to its own training
length. To avoid breaking comparability the setting was not changed, and this
mismatch is treated as a lower bound on the results.

\paragraph{Metrics.}
The benchmark's own metric set is used: nDCG@10 \citep{jarvelin2002ndcg},
nDCG@100, Recall@100 and mAP.
The deep cutoff (@100) is definitional for late-interaction models their
typical deployment is the reranking of a deep candidate list, and @10 measures
only the top of that capability.

\paragraph{\maxsim{} correctness test.}
An axis or mask-orientation error in a \maxsim{} implementation produces silently
wrong but plausible-looking scores the ranking breaks while the score
distribution still looks normal. To catch this, a unit test in which a known
positive must rank above known negatives is executed before the full evaluation.

% =====================================================================
\section{Results}
\label{sec:results}

\subsection{Overall result}

Table~\ref{tab:average} reports the averages over the five datasets. \model{}
finishes second on all four metrics and outperforms all three of the other
late-interaction models compared. Two numbers stand out:

\begin{itemize}\setlength{\itemsep}{0.15em}
\item \textbf{Half the size, higher score.} \model{} beats the twice-as-large
      ColmmBERT-base-TR by $+3.05$ overall, $+2.60$ nDCG@10, $+2.99$ nDCG@100,
      $+3.79$ Recall@100 and $+2.84$ mAP. The largest model in the benchmark,
      LFM2.5-ColBERT-350M, is $12.30$ points behind overall. The relation
      between parameter count and score is therefore weak in this benchmark;
      what matters is the language fit of the encoder.
\item \textbf{Clear first in its own size class.} The nearest model by size is
      ColmmBERT-small-TR, and the gap is $+4.59$ overall and $+5.71$
      Recall@100.
\end{itemize}

The leading model, mLateOn, is ahead on every metric; the gap is $5.27$ points
overall and $6.61$ on Recall@100. How that gap is distributed is the subject of
\S\ref{sec:discussion}.

\begin{restable}
\centering
\caption{Turkish BEIR retrieval performance, averaged over five datasets. All
models were evaluated with the official \TurkColBERT{} pipeline (PLAID index,
\code{document\_length}=300, $k$=100). \textbf{OVERALL} is the mean of the four
metrics. Best per column in bold.}
\label{tab:average}
\small
\begin{tabular}{lrrrrrc}
\toprule
Model & Params & nDCG@10 & nDCG@100 & R@100 & mAP & \textbf{OVERALL} \\
\midrule
mLateOn \citep{sourty2026lateon} & 306.9M & \best{37.23} & \best{40.72} & \best{63.59} & \best{28.98} & \best{42.63} \\
\ours{\model{}} & 148.9M & 31.81 & 35.53 & 56.98 & 25.13 & \ours{37.36} \\
ColmmBERT-base-TR \citep{ezerceli2025turkcolbert} & 306.9M & 29.21 & 32.54 & 53.19 & 22.29 & 34.31 \\
ColmmBERT-small-TR \citep{ezerceli2025turkcolbert} & 140.5M & 27.67 & 30.96 & 51.27 & 21.16 & 32.77 \\
LFM2.5-ColBERT-350M \citep{liquidai2026colbert} & 353.3M & 20.28 & 23.27 & 41.85 & 14.83 & 25.06 \\
\bottomrule
\end{tabular}
\end{restable}

\subsection{Per-dataset results}

Table~\ref{tab:perdataset} reports all five datasets separately. The pattern is
consistent with a single exception: \model{} is second on four datasets and
fourth on one.

\begin{restable}
\centering
\caption{Per-dataset retrieval performance; within each block models are ordered
by overall score. \textbf{OVERALL} is the mean of the four metrics. Best per
column in bold. For parameter counts see Table~\ref{tab:average}.}
\label{tab:perdataset}
\small
\begin{tabular}{lrrrrc}
\toprule
Model & nDCG@10 & nDCG@100 & R@100 & mAP & \textbf{OVERALL} \\
\midrule
\multicolumn{6}{l}{\textit{SciFact-TR} scientific claim verification}\\
\midrule
mLateOn & \best{71.22} & \best{74.12} & \best{94.50} & \best{68.21} & \best{77.01} \\
\ours{\model{}} & 66.89 & 70.20 & 90.43 & 64.14 & \ours{72.92} \\
ColmmBERT-base-TR & 59.65 & 63.00 & 85.36 & 56.51 & 66.13 \\
ColmmBERT-small-TR & 58.77 & 61.91 & 84.72 & 55.21 & 65.15 \\
LFM2.5-ColBERT-350M & 44.89 & 48.96 & 78.79 & 40.34 & 53.25 \\
\midrule
\multicolumn{6}{l}{\textit{ArguAna-TR} --- argument retrieval}\\
\midrule
mLateOn & \best{38.85} & \best{43.15} & \best{97.65} & \best{27.10} & \best{51.69} \\
\ours{\model{}} & 30.39 & 36.17 & 88.76 & 21.46 & \ours{44.20} \\
ColmmBERT-base-TR & 24.42 & 30.25 & 78.59 & 17.21 & 37.62 \\
ColmmBERT-small-TR & 22.37 & 28.57 & 75.96 & 15.98 & 35.72 \\
LFM2.5-ColBERT-350M & 16.02 & 21.47 & 60.24 & 11.45 & 27.30 \\
\midrule
\multicolumn{6}{l}{\textit{FiQA-TR} --- financial answer retrieval}\\
\midrule
mLateOn & \best{33.72} & \best{40.34} & \best{66.47} & \best{27.89} & \best{42.11} \\
\ours{\model{}} & 26.90 & 33.09 & 55.53 & 22.27 & \ours{34.45} \\
ColmmBERT-base-TR & 24.35 & 30.14 & 52.47 & 19.55 & 31.63 \\
ColmmBERT-small-TR & 21.33 & 26.81 & 47.96 & 17.08 & 28.30 \\
LFM2.5-ColBERT-350M & 10.42 & 14.22 & 28.99 &  8.04 & 15.42 \\
\midrule
\multicolumn{6}{l}{\textit{SciDocs-TR} --- citation prediction}\\
\midrule
mLateOn & \best{13.61} & \best{20.03} & \best{33.34} & \best{9.09} & \best{19.02} \\
\ours{\model{}} & 11.64 & 17.02 & 28.33 & 7.68 & \ours{16.17} \\
ColmmBERT-base-TR & 10.10 & 15.52 & 26.59 & 6.73 & 14.74 \\
ColmmBERT-small-TR &  9.42 & 14.22 & 24.36 & 6.16 & 13.54 \\
LFM2.5-ColBERT-350M &  7.01 & 11.48 & 21.03 & 4.51 & 11.01 \\
\midrule
\multicolumn{6}{l}{\textit{NFCorpus-TR} --- biomedical/nutrition}\\
\midrule
mLateOn & \best{28.75} & \best{25.95} & \best{26.00} & \best{12.60} & \best{23.33} \\
ColmmBERT-base-TR & 27.56 & 23.80 & 22.93 & 11.44 & 21.43 \\
ColmmBERT-small-TR & 26.45 & 23.27 & 23.32 & 11.36 & 21.10 \\
\ours{\model{}} & 23.21 & 21.18 & 21.83 & 10.08 & \ours{19.08} \\
LFM2.5-ColBERT-350M & 23.06 & 20.22 & 20.17 &  9.79 & 18.31 \\
\bottomrule
\end{tabular}
\end{restable}

\paragraph{Where the gains concentrate.}
The nDCG@10 difference against ColmmBERT-base-TR is not evenly distributed across
datasets: SciFact-TR $+7.24$, ArguAna-TR $+5.97$, FiQA-TR $+2.55$, SciDocs-TR
$+1.54$, NFCorpus-TR $-4.35$. The two largest gains occur on the two datasets
where the query is not a keyword phrase but a \emph{full proposition}: in SciFact
the query is a scientific claim, in ArguAna an entire argument. Both tasks
require locating which sentence of a document supports the claim exactly the
kind of operation late interaction is designed to be strong at.

\paragraph{The single loss: NFCorpus-TR.}
\model{} falls behind both Turkish ColmmBERT models on NFCorpus-TR and finishes
only $0.15$ points ahead of LFM2.5. This dataset is in the nutrition and
biomedical domain; the documents are PubMed abstracts and most of the
discriminative signal is concentrated in specialist terminology drug,
nutrient and disease names. The composition of our pretraining corpus is weighted
towards web, legal and general academic text, and the biomedical domain was not
additionally upweighted. This repeats exactly the pattern observed for the
single-vector model in the earlier work: we lead on tasks requiring structural
and semantic parsing, and fall behind on tasks decided by the breadth of domain
vocabulary.

% =====================================================================
\section{Discussion}
\label{sec:discussion}

\paragraph{A monolingual encoder matters more than parameters.}
The ordering in Table~\ref{tab:average} does not track parameter count: the
largest model is last, and the second-placed model is the second smallest in the
list. Most late-interaction entries in the benchmark are Turkish adaptations of a
multilingual or English base the other two models in the top three also
descend from mmBERT \citep{marone2025mmbert}; \model{} instead comes from an entirely Turkish corpus and a 50,048-piece
tokenizer trained for Turkish. We believe tokenizer fit feeds late interaction
directly in an agglutinative language: with 95\% of word types split into
multiple pieces, whether those pieces are cut at morphologically meaningful
boundaries determines the quality of the units \maxsim{} matches. This is a hypothesis and should be tested with an ablation that holds the
tokenizer fixed and varies the encoder. Its direction is consistent with the
JaColBERT results measuring the same contrast in Japanese
\citep{clavie2024jacolbertv25}; the opposing strategy modular transfer from
a single high-resource language is advocated by ColBERT-XM
\citep{louis2024colbertxm}, and the two have not been compared directly on
Turkish.

\paragraph{Where the gap to the leader concentrates.}
The nDCG@10 deficit against mLateOn ranges from $1.97$ (SciDocs) to $8.46$
(ArguAna). The largest deficit is on ArguAna and is even more pronounced on
Recall@100 ($88.76$ vs.\ $97.65$, $-8.89$). ArguAna has by far the longest
queries among the five datasets the query is not a keyword phrase but a full
argument paragraph. \model{}'s query encoder is fixed at 32 tokens
(Table~\ref{tab:config}); a large part of such a query therefore never reaches
the model. This does not explain the entire deficit, but the largest deficit
appearing on the dataset with the longest queries cannot be read as coincidence.
We derive a testable prediction: if \code{query\_length} is raised to 128 or 256
and the model retrained, the gain should concentrate disproportionately on
ArguAna.

\paragraph{The document window is below the training length.}
The official pipeline uses \code{document\_length}=300 while the model was
trained with 512. Since the entire rationale for the long-band choice
(250--2500 tokens) rests on a 512-token window, the reported numbers were
measured in a setting below the regime the model was trained for. We did not
change the setting for the sake of comparability; the results should therefore be
read as a lower bound rather than an upper one.

\paragraph{The role of distillation.}
Training is a single epoch and a single phase; the multi-round, index-guided
negative refresh loop of ColBERTv2 \citep{santhanam2022colbertv2} was not
applied. Even so, the resulting ranking shows that cross-encoder distillation
combined with a sufficiently clean negative pool yields a competitive
late-interaction model in one pass; the same observation is reported for
JaColBERTv2.5 \citep{clavie2024jacolbertv25} under a constrained budget.
The cleanliness of the negative pool is the real variable here: without the group
mask of \S\ref{sec:negatives}, the second passage carved from the same document
would systematically enter training labelled as a negative.

\paragraph{Deployment.}
A multi-vector index is tens of times larger than a dense one, depending on
document length. The recommended setup in practice is two-stage candidate
retrieval with a single-vector model, reranking with \model{}. Standalone ColBERT
retrieval is possible, and the measurements in this work were made in exactly
that regime (PLAID); but as the corpus grows, a compression or pruning layer such as PLAID
\citep{santhanam2022plaid}, XTR \citep{lee2023xtr} or MUVERA
\citep{dhulipala2024muvera} becomes a practical necessity.

% =====================================================================
\section{Limitations}
\label{sec:limitations}

\begin{itemize}\setlength{\itemsep}{0.2em}
\item \textbf{One epoch, one seed.} The model was trained with a single epoch and
      a single seed; the reported differences were not separated from
      within-run noise.
\item \textbf{No ablations.} Distillation vs.\ contrastive loss, \dense{} vs.\ a
      raw encoder as initialization, the passage cap, the cosine ceiling and the
      query length were not measured; all are justified but empirically
      unvalidated design decisions.
\item \textbf{Query length 32.} As discussed in \S\ref{sec:discussion}, this
      ceiling is a source of loss on long-query tasks and was not measured.
\item \textbf{Evaluation window below the training window.} The official
      pipeline uses \code{document\_length}=300; no measurement was made at 512.
\item \textbf{Five datasets, one benchmark.} Although the datasets cover five
      distinct retrieval types, the benchmark's translation quality and qrels
      decisions are a shared source of systematic bias; no validation was
      performed on an independent Turkish retrieval collection.
\item \textbf{The MUVERA path was not measured.} Only PLAID (exact \maxsim) was
      run; the model's speed/quality curve under approximate search is unknown.
\item \textbf{Domain coverage.} The NFCorpus-TR result indicates that the
      pretraining corpus is weak in the biomedical domain; this is a limitation
      of corpus composition and is independent of the architecture.
\item \textbf{A single teacher.} One cross-encoder was used; the teacher's own
      biases on Turkish may have been inherited by the student through
      distillation.
\end{itemize}

% =====================================================================
\section{Conclusion}

We introduced the first Turkish late-interaction retrieval model derived from a
ModernBERT encoder trained from scratch on an entirely Turkish corpus. \model{}
is initialized from the encoder of our dense embedding model, adapted to the
ColBERT objective with a token-level $768\!\rightarrow\!128$ projection, and
trained with a single epoch of cross-encoder distillation. Its training data is
produced from our own pretraining corpus without decoding and by splitting at
sentence boundaries; the hard negative pool is cleaned with a group mask and a
cosine ceiling on top of rank-based skipping.

Measured on \TurkColBERT{} with the benchmark's official pipeline, and with all
five datasets clean zero-shot, \model{} finishes second among five models with an
overall score of 37.36: it outperforms the twice-as-large ColmmBERT-base-TR on
four of five datasets and by $+3.05$ overall, and the benchmark's largest model
LFM2.5-ColBERT-350M by $+12.30$. This shows that what is decisive
is not parameter count but the language fit of the encoder. The model's only
regression is on NFCorpus-TR in the biomedical domain, explained by the domain
composition of the pretraining corpus; its largest deficit against the leading
model is on ArguAna-TR, pointing to the 32-token ceiling of the query encoder.

Two immediate next steps follow from those two observations: raising the query
length ceiling and retraining, and adding Turkish biomedical text to either the
pretraining or the ColBERT phase. Both are measurable and directly comparable
under the protocol established in this paper.

% =====================================================================

% =====================================================================

\end{document}